\pdfoutput=1

\documentclass[11pt]{article}

\usepackage{EMNLP2022}
\usepackage{makecell}
\usepackage{times}
\usepackage{latexsym}
\usepackage{amsmath}
\usepackage{tikz}
\usepackage{pgfplots}
\usepackage[T1]{fontenc}
\usepackage[utf8]{inputenc}

\usepackage{microtype}

\usepackage{inconsolata}
\usepackage{booktabs}
\usepackage{multirow}
\usepackage{dsfont}
\usepackage{arydshln}
\usepackage{enumitem}
\usepackage{graphics}
\usepackage{graphicx}
\usepackage{subfigure}
\usepgfplotslibrary{colorbrewer}
\usepackage{color}
\usetikzlibrary{arrows.meta,calc,decorations.markings,math,arrows.meta,patterns}
\usetikzlibrary{decorations.pathreplacing}
\usetikzlibrary{matrix}

\definecolor{lizhaocong}{rgb}{1,0,0}
\definecolor{lizhaocong_camera}{rgb}{0,0,1}
\newcommand{\lzc}[1]{{\color{black} #1}}
\newcommand{\derek}[1]{{\color{black} #1}}
\title{ConsistTL: Modeling Consistency in Transfer Learning \\ for Low-Resource Neural Machine Translation}

\author{Zhaocong Li$^1$~~~
        Xuebo Liu$^2$\thanks{~~Co-corresponding author}~~~
        Derek F. Wong$^1\footnotemark[1]$~~~
        Lidia S. Chao$^1$~~~
        Min Zhang$^2$\\
  $^1$NLP$^2$CT Lab, Department of Computer and Information Science, 
  University of Macau \\
      \texttt{nlp2ct.zhaocong@gmail.com, \{derekfw,lidiasc\}@um.edu.mo} \\
    $^2$Institute of Computing and Intelligence, Harbin Institute of Technology, Shenzhen, China \\
          \texttt{\{liuxuebo,zhangmin2021\}@hit.edu.cn} }

\begin{document}
\maketitle
\begin{abstract}
Transfer learning is a simple and powerful method that can be used to boost model performance of low-resource neural machine translation (NMT).
Existing transfer learning methods for NMT are static, which simply transfer knowledge from a parent model to a child model once via parameter initialization.
In this paper, we propose a novel transfer learning method for NMT, namely ConsistTL, which can continuously transfer knowledge from the parent model during the training of the child model.
Specifically, for each training instance of the child model, ConsistTL constructs the semantically-equivalent instance for the parent model and encourages prediction consistency between the parent and child for this instance, which is equivalent to the child model learning each instance under the guidance of the parent model. 
Experimental results on five low-resource NMT tasks demonstrate that ConsistTL results in significant improvements over strong transfer learning baselines, with a gain up to 1.7 BLEU over the existing back-translation model on the widely-used WMT17 Turkish-English benchmark.
Further analysis reveals that ConsistTL can improve the inference calibration of the child model.
Code and scripts are freely available at \url{https://github.com/NLP2CT/ConsistTL}.

\end{abstract}

\section{Introduction}
Neural machine translation (NMT) has achieved success on the high-resource language pairs \citep{vaswani2017attention}. 
However, it achieves an inadequate performance on the low-resource language pairs with merely bilingual data due to the data sparsity \citep{DBLP:conf/aclnmt/KoehnK17,DBLP:conf/acl/SennrichZ19}. 
Transfer learning is a simple and powerful method that can be used to boost the model performance of low-resource NMT, which can transfer knowledge from an off-the-shelf high-resource parent model to the low-resource child model \citep{zoph-etal-2016-transfer}.

\begin{figure}[t]
    \centering
    \subfigure[Vanilla TL]{\label{fig:Standard}\includegraphics[scale=0.13]{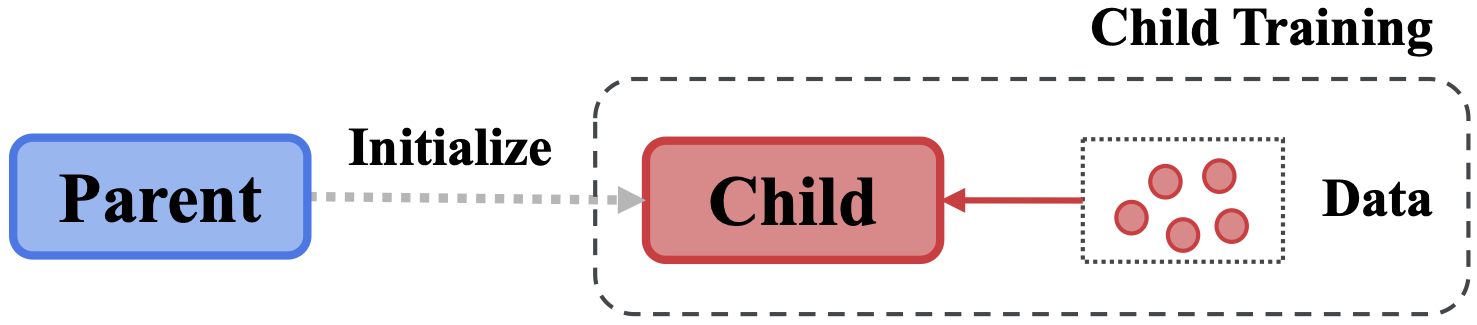}}
    \hfill
    \subfigure[ConsistTL]{\label{fig:ConsistTL}\includegraphics[scale=0.13]{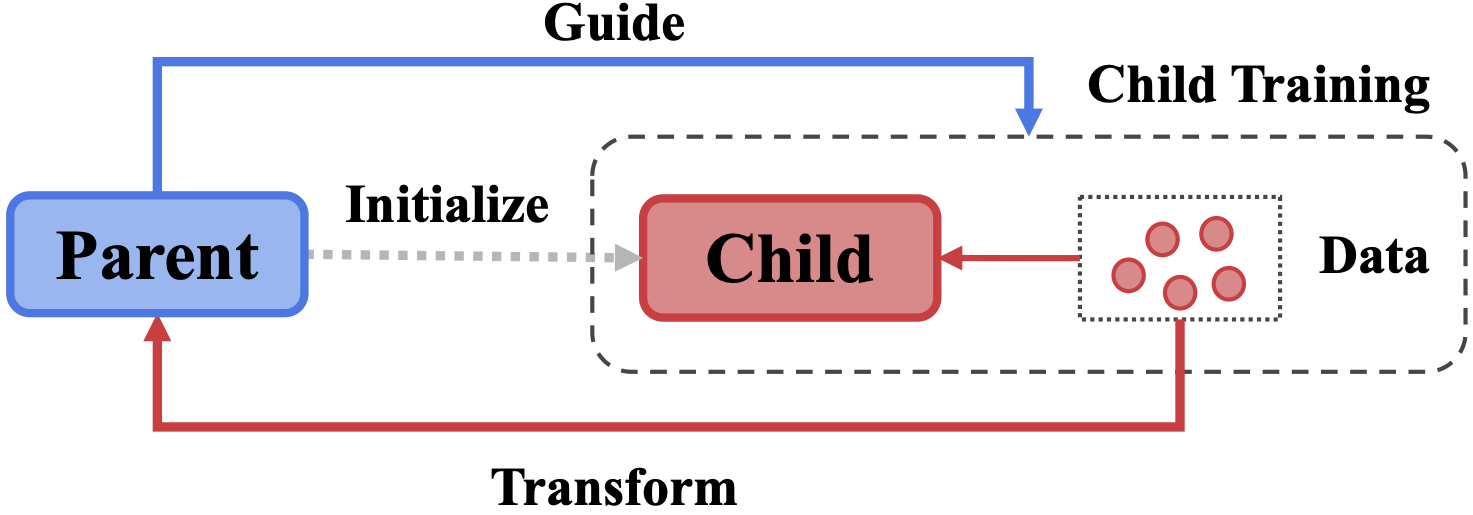}}
    \caption{Comparison between vanilla TL and ConsistTL. The dashed arrow represents the initialization of the parameters from the parent model. Apart from disposable parameter transfer, ConsistTL continuously transfers knowledge from the parent model during the child training.}
    \label{fig:comparison}
\end{figure}

The goal of transfer learning for low-resource NMT is to transfer knowledge sufficiently from the parent model. 
Prior works attempt to transfer more information from the embedding layer of the parent model, using methods such as building a joint dictionary \citep{DBLP:conf/wmt/KocmiB18,DBLP:journals/corr/abs-1909-06516}, creating cross-lingual token mapping \citep{DBLP:conf/acl/KimGN19}, and transferring the word of same form  \citep{DBLP:conf/acl/AjiBHS20}.
These strategies achieve success in parameter transfer. 
The methods used in these works is summarized in Figure \ref{fig:Standard}: the parent model transfers its parameters to the child model once and lacks continual guidance on the child training. This method can not utilize the parent model sufficiently, which potentially limits the learning of the child model.

This paper attempts to leverage a continual knowledge transfer from the parent model for the child training. {The prediction distribution of the parent model is usually more informative since it has learned to predict various translations during training}~\citep{dreyer-marcu-2012-hyter,DBLP:conf/emnlp/KhayrallahTPK20}. 
Therefore, utilizing the prediction of the parent model to continuously guide the child training might be a nice plus.

To this end, we propose \textbf{Consist}ency-based \textbf{T}ransfer \textbf{L}earning (\textbf{ConsistTL}), a novel framework that can be used to model the consistency between the parent model and the child model during the child training. 
For each instance from the child data, ConsistTL constructs a semantically-equivalent pseudo source sentence for the parent by translating the child target sentence. 
{Then ConsistTL encourages the child prediction to be consistent with the parent prediction since these two predictions are conditioned on the same target sentence and semantically-equivalent source sentences.} 
ConsistTL is summarized in Figure \ref{fig:ConsistTL}: unlike in vanilla TL, the parent model releases continual guidance for the child model during training by receiving the transformation of the training data of the child model. 
With ConsistTL, the child model can acquire more knowledge from the parent predictions beyond the simple parameter transfer. 
We conduct experiments on five low-resource NMT benchmarks. Experimental results show that ConsistTL achieves significant improvements over strong baselines of transfer learning. 
Combined with the representative data augmentation method of back-translation~\cite{DBLP:conf/acl/SennrichHB16}, our method can outperform the existing method up to 1.7 BLEU on the widely-used WMT17 Turkish-English benchmark. An extensive analysis reveals that our method, as an approach to model external consistency, is complementary to the method of modeling inner consistency.

{
Our contributions are as follows:
\begin{itemize}
    \item We propose ConsistTL to model the consistency between the parent and child models in transfer learning for low-resource NMT. The cross-model consistency extends the knowledge transfer process to child training.
    \item ConsistTL outperforms strong baselines significantly on five low-resource benchmarks, and is complementary to the other methods for low-resource NMT (e.g., back-translation and inner-model consistency learning).
    \item We find that ConsistTL can improve the calibration of the child model by increasing translation accuracy and reducing overconfidence.
\end{itemize}}

\section{Related Work}
\paragraph{Transfer Learning for NMT} \citet{zoph-etal-2016-transfer} propose a parent-child framework to transfer the parameters from the parent model trained on a high-resource corpus to a low-resource child model with the same target language. 
The follow-up works aim to transfer the knowledge from the parent model parameters more sufficiently since \citet{zoph-etal-2016-transfer} ignore the embedding layer mismatch caused by the vocabulary mismatch. 
These works include building the shared joint dictionary \citep{DBLP:conf/wmt/KocmiB18,DBLP:journals/corr/abs-1909-06516,liu2019latent,liu-etal-2019-shared}, using an extra trained transformation to connect the embedding layers of two models \citep{DBLP:conf/acl/KimGN19,sato2020vocabulary,liu2021bridging}, or transferring partial word embedding from the parent model \citep{DBLP:conf/acl/AjiBHS20,DBLP:conf/acl/XuH22}. 
In comparison, this presenting paper tries to transfer the knowledge contained in the parent prediction beyond parameters.

\begin{figure*}[t]
    \centering
    \includegraphics[width=1\textwidth]{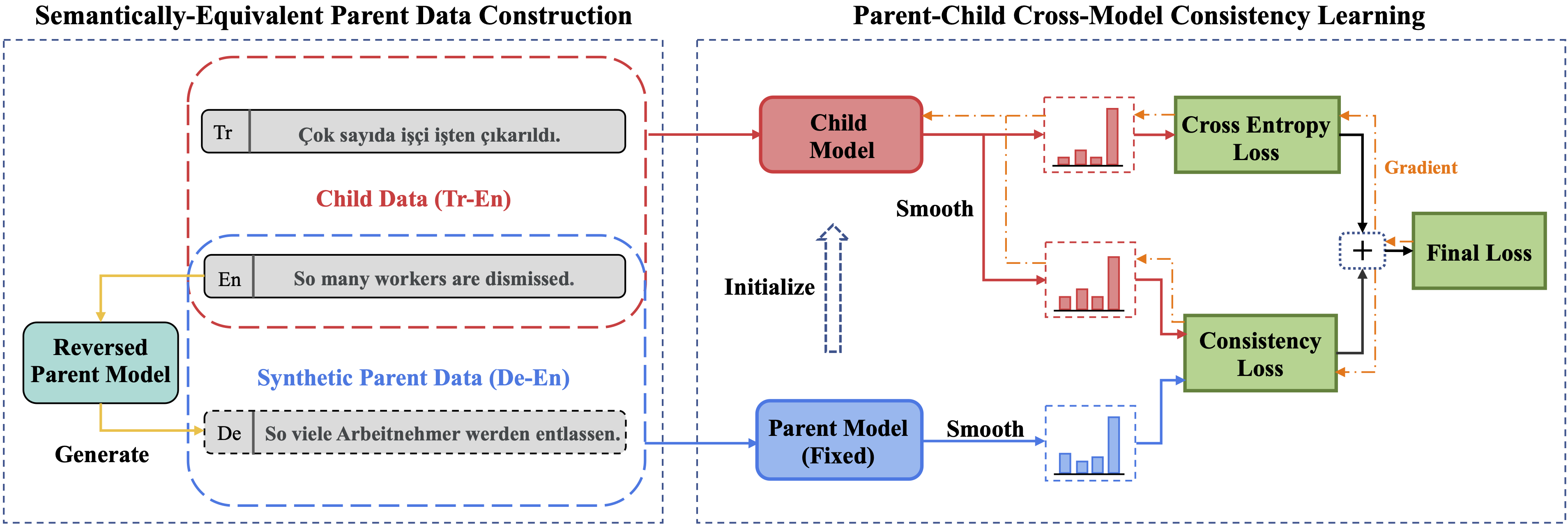}
    \caption{Overview of ConsistTL. ConsistTL first constructs semantically-equivalent pseudo parent source sentences from child target sentences using the reversed parent model. Then, given the sentence group with the same semantic meaning, ConsistTL encourages the child prediction to be consistent with the parent prediction during child training.}
    \label{fig:framework}
\end{figure*}
There are also some works that consider multilingualism to enhance transfer learning. 
\citet{DBLP:conf/emnlp/NeubigH18,tan2019study} propose to transfer knowledge from a multilingual parent model. \citet{gu-etal-2018-universal} develop a universal architecture with shared lexical and sentence level representation to implement transfer learning. 
\citet{gu-etal-2018-meta} propose a meta-learning approach to learn the initialization process using several high-resource language pairs. In this paper, we mainly focus on the bilingual parent model, since off-the-shelf bilingual models are easier to obtain.

\paragraph{Consistency Learning in NMT}
Consistency learning aims to model consistency across different model predictions. The core idea is to craft the cross-view supervision and enhance cross-view consistency, which enriches the supervision signals. In the early studies, consistency learning methods focus on semi-supervised learning since consistency signals enable models to learn from data without manual labels \citep{DBLP:conf/nips/XieDHL020,DBLP:journals/pami/MiyatoMKI19,DBLP:conf/emnlp/LowellHLW21}. 
The semi-supervised consistency learning also enhances NMT models with the combination of forward-translation \citep{DBLP:conf/emnlp/ClarkLML18}. 
Recent studies of consistency learning gradually consider supervised learning areas \citep{chen-etal-2021-hiddencut, lowell-etal-2021-unsupervised}. 
R-Drop is one of the simple and generalized methods~\citep{DBLP:conf/nips/LiangWLWMQCZL21}, which enhances the consistency between multiple model structures produced by dropout, and improves NMT performance.

There are some consistency learning methods specially designed for supervised NMT based on data augmentation.
\citet{shen2020simple, DBLP:conf/acl/GuoMZ022,gao-etal-2022-bi} leverage token-level perturbations to craft different predictions. \citet{DBLP:conf/acl/KambhatlaBS22} explore the insight of cryptology to create a ciphertext with the same semantic meaning for each training instance. 
\citet{xie2021target} develop a scheme to augment target side input by the output prediction from the first forward. 
\citet{stroke2022} enhance the Chinese representation by modeling the consistency between stroke representation and its corresponding encryption. 
These consistency modeling works in NMT mainly exploit the inner model consistency, i.e., modeling consistency within the same model. Compared with these works, we exploit transferable information from the cross-model consistency in transfer learning for low-resource NMT.
\citet{chen-etal-2017-teacher} also exploit the consistency between two models and the differences from us are: 1) we construct the parent data instead of the child data; 2) we enhance the transfer learning framework for low-resource NMT while their method relies on a pivot language to improve zero-resource NMT.

\section{ConsistTL}
\subsection{Motivation}
Previous works concentrate on knowledge transfer from the parent model parameters \citep{DBLP:conf/wmt/KocmiB18,DBLP:conf/acl/KimGN19,DBLP:conf/acl/AjiBHS20}. 
This paper argues that continual knowledge transfer from the predictions of the parent model is also beneficial to transfer learning. 
Since the parent model is trained on a large number of diverse translations~\citep{dreyer-marcu-2012-hyter,DBLP:conf/emnlp/KhayrallahTPK20}, the prediction of the parent model might be informative for child training. 

In this paper, we propose ConsistTL, a framework that can be used to jointly transfer the knowledge from the parent parameters and the prediction distribution of the parent model. 
ConsistTL can enhance the consistency between the distribution from the parent model (e.g., German-English) and the child model (e.g., Turkish-English) conditioned on the semantically-equivalent sentences. ConsistTL is composed of two parts: semantically-equivalent parent data construction and parent-child cross-model consistency learning. 
The overall framework is illustrated in Figure \ref{fig:framework}.

\subsection{Semantically-Equivalent Parent Data Construction}
To implement the proposed ConsistTL, we need to model the cross-model consistency between the parent-side source sentence and child-side source sentence.
As the parent model and child model are trained for different language pairs, for each instance in the child data, we need a semantically-equivalent sentence written in the source language of the parent language pair.

However, during the child training, we may have only the small-scale bilingual child data $\mathcal{D}_c = \{\langle\boldsymbol{x}_c^n,\boldsymbol{y}^n\rangle\}_{n=1}^N$. Acquiring a semantically-equivalent source sentence for each instance of the parent in $\mathcal{D}_c$ is labor-consuming. 
Inspired by back-translation \citep{DBLP:conf/acl/SennrichHB16}, ConsistTL creates synthetic parent data by generating pseudo source sentences $\Tilde{\boldsymbol{x}}_p$ from $\boldsymbol{x}_c$, or $\boldsymbol{y}$.

The translation quality is crucial for generating pseudo sentences since low-quality translation would cause poor semantic equivalence between the generated sentence $\Tilde{\boldsymbol{x}}_p$ and the existing sentence pair $\langle\boldsymbol{x}_c, \boldsymbol{y}\rangle$. 
Obviously, the reversed parent model $\mathcal{M}_{\boldsymbol{y}\xrightarrow{}\boldsymbol{x_p}}$ is easier to obtain and tends to produce higher quality translation than the low-resource translation model $\mathcal{M}_{\boldsymbol{x_c}\xrightarrow{}\boldsymbol{x_p}}$. 
Therefore, we choose to back-translate the target sentence $\boldsymbol{y}$ into $\Tilde{\boldsymbol{x}}_p$ using the reversed parent model $\mathcal{M}_{\boldsymbol{y}\xrightarrow{}\boldsymbol{x_p}}$ in ConsistTL. Then, we can concatenate the pseudo source data $\{\Tilde{\boldsymbol{x}}_p^n\}_{n=1}^N$ with the target side data $\{\boldsymbol{y}^n\}_{n=1}^N$ from the child language pair resulting in the synthetic parent data $\Tilde{\mathcal{D}_p} = \{\langle\Tilde{\boldsymbol{x}}_p^n, \boldsymbol{y}^n\rangle\}_{n=1}^N$. 
After constructing the synthetic parent data, ConsistTL steps into the next stage to take advantage of this synthetic parent data.
\subsection{Parent-Child Cross-Model Consistency Learning}
In this part, we encourage the child model to learn the cross-model consistency information from the child data and synthetic parent data. 
{Every instance $\langle\boldsymbol{x}_c,\boldsymbol{y}\rangle$ from the child data $\mathcal{D}_c$ and its semantically-equivalent counterpart $\langle\Tilde{\boldsymbol{x}}_p,\boldsymbol{y}\rangle$ from the synthetic parent data $\Tilde{\mathcal{D}_p}$ are fed into the child and parent models, respectively, at the same time. 
The corresponding prediction distributions can be obtained from their forwards.}
Our goal is to make these two distributions as consistent as possible. 
We propose a cross-model consistency loss to make the parent model distribution and the child model distribution closer during child training.

\paragraph{Cross-Model Consistency Loss} To make the distribution of the parent model and child model consistent, we need to measure and minimize the gap between these distributions. %
The proposed framework minimizes the distribution gap with the following cross-model loss:
\begin{equation}
\begin{aligned}
    \mathcal{L}_d(\theta_c) =\sum_{t=1}^{T}
    F[P(\boldsymbol{*}|\boldsymbol{y}_{<t}, \boldsymbol{x}_c, \theta_c, \tau),\\
    P(\boldsymbol{*}|\boldsymbol{y}_{<t},\Tilde{\boldsymbol{x}}_p, \theta_p^*,\tau)]
\end{aligned}
\label{eq}
\end{equation}
where $F$ measures the distribution gap between parent model and child model. We choose Jensen–Shannon divergence \citep{DBLP:journals/tit/Lin91} as $F$. $\theta_p^*$ and $\theta_c$ represent parameters of parent model and child model respectively. 
$P(\boldsymbol{*}|\boldsymbol{y}_{<t},\Tilde{\boldsymbol{x}}_p, \theta_p^*, \tau)$ and $P(\boldsymbol{*}|\boldsymbol{y}_{<t}, \boldsymbol{x}_c, \theta_c, \tau)$ denote the prediction distributions of parent model and child model at time step $t$ respectively. 
The parent parameters are fixed during child training as it is well trained on the parent language pair. 
We use $\tau$ to control the smoothness of the prediction distribution:
\begin{equation}
    \begin{aligned}
    P(y_t=k|\boldsymbol{y}_{<t}, \boldsymbol{x}, \theta) = \frac{exp(z_k/\tau)}{\sum_{i=1}^{|V|}exp(z_i/\tau)}
    \end{aligned}
\end{equation}
where $z_i$ denotes the logits output before softmax is computed, and $|V|$ represents the size of the target-side vocabulary. 
We can smooth the distribution of the acquired prediction with $\tau > 1$. 

\paragraph{Final Child Training Objective Function} We interpolate the weighted cross-model consistency loss and negative log-likelihood loss to create the final child training objective function. The final child training objective is formulated as:
\begin{equation}
    \mathcal{L}(\theta_c) = \mathcal{L}_{nll}(\theta_c) + \alpha\mathcal{L}_d(\theta_c)
\end{equation}
In this manner, the child model can learn each instance under the guidance of the parent model.

\section{Experiments}
\subsection{Datasets}
\paragraph{Parent Language Pair} {This paper follows~\citet{DBLP:conf/acl/AjiBHS20} to transfer knowledge from the German-English (De-En) task.} Our parent model is trained on the WMT17 {De-En} training set and validated on \textit{newstest2013}. The training set consists of 5.8M sentences. The vocabulary is implemented \derek{using} joint source-target BPE with 40k merge operations \citep{DBLP:conf/acl/SennrichHB16a}.
\paragraph{Child Language Pairs} We conduct transfer learning experiments on five low-resource translation benchmarks. There are four translation benchmarks from Global Voices \citep{DBLP:conf/lrec/Tiedemann12,DBLP:conf/emnlp/KhayrallahTPK20}: Polish (Pl), Hungarian (Hu), Indonesian (Id), and Catalan (Ca). 
{The sentences of corresponding languages are paired with English sentences.} 
The subset splits follow \citet{DBLP:conf/emnlp/KhayrallahTPK20}. 
For the fifth language pair, we adopt WMT17 Turkish (Tr)-English benchmark and use \textit{newstest2016} as the validation set. Before subword segmentation, we apply normalization and tokenization to all datasets \derek{using the} Moses~\citep{koehn-etal-2007-moses}. The sentences with more than 60 words and \derek{a length ratio of over 1.5} in training data would be filtered for the Tr-En language pair. After data filtering, we then apply joint source-target BPE \derek{to the} child language pairs with fewer merge operations than \derek{were used for the} parent language pair.
We display detailed statistics of \derek{the} preprocessed datasets in Table \ref{tab:data}. 
\begin{table}[t]
    \centering
    \begin{tabular}{lrrrr}
    \toprule
    \textbf{Task} & \textbf{Train} & \textbf{Valid} &  \textbf{Test} & \textbf{BPE}\\
    \midrule
      Tr-En   & 192,614 & 3,000 & 3,007 & 16,000 \\
      \hdashline
      Pl-En & 39,943 & 2,000 & 2,000 & 8,000 \\
      Ca-En & 15,176 & 2,000 & 2,000 & 8,000 \\
      Id-En & 8,448 & 2,000 & 2,000 & 8,000 \\
      Hu-En & 7,712 & 2,000 & 2,000 & 8,000 \\
      \bottomrule
    \end{tabular}
    \caption{Numbers of sentences and BPE merge operations of preprocessed datasets.}
    \label{tab:data}
\end{table}
\subsection{Model Configuration}
We use \derek{the} Transformer-base architecture \citep{vaswani2017attention} to implement NMT models in our experiments from the toolkit \textsc{Fairseq} \citep{ott2019fairseq}. For all the low-resource translation models, we tie the input embedding layers of \derek{the} decoder and \derek{the} output projections \citep{DBLP:conf/eacl/PressW17, DBLP:conf/iclr/InanKS17}. For the high-resource parent model, we tie all embedding layers.
\paragraph{Vanilla NMT} We implement the transformer models \lzc{training from scratch as baselines}. These models share the vocabularies of source and target languages.
\paragraph{TL} \citet{zoph-etal-2016-transfer} propose a transfer learning method to randomly \lzc{assign a parent word embedding to each child word}. The rest of the parameters are copied from the parent model. We implement this method as a baseline in our experiment. This vanilla transfer learning method is marked as ``TL''.
\paragraph{TM-TL} {\citet{DBLP:conf/acl/AjiBHS20} propose a method named ``token matching'' (TM), which is simple and effective. We implement this transfer learning method as a baseline in our experiments marked as ``TM-TL''. Some tokens are common to both parent model and child model vocabularies for child language. TM-TL assigns the embedding parameters of such common tokens in the source embedding layer of the child model. For the other parameters of the source embedding layer, TM-TL initializes them randomly following the vanilla NMT. The rest settings of TM-TL follow TL.}
\paragraph{ConsistTL} Since our method\derek{,} ConsistTL\derek{,} transfers knowledge from the prediction distribution of \derek{the} parent model, it is orthogonal to the parameter transfer methods. We adopt TM to implement ConsistTL.

\subsection{Settings}
\label{training}
\paragraph{Training} We train all the NMT models using \derek{the} Adam optimizer \citep{DBLP:journals/corr/KingmaB14} with $\beta = (0.9, 0.98)$. We also use the inverse square root schedule to control \derek{the variation of the} learning rate during training. For \derek{the} parent model, the value of linear warmup steps is set \derek{to} $10,000$ and the peak learning rate is set \derek{to} $0.001$. The parent model is trained for 200 epochs with 460K ($3584 \times 128$) tokens per batch and a low dropout rate of 0.1. We use 4,000/2,000 max tokens per batch for Tr/Pl-En tasks, and 1,000 for the other tasks. For the \derek{the training of the} low-resource translation model from scratch, we set the \derek{number} of linear warm-up steps \derek{to} $8,000$ and the peak learning rate \derek{is set to} $0.0005$. For the low-resource translation models with transfer learning, as \derek{the} inner layers are adequately trained on \derek{the} parent language pair, we set narrow warm-up steps \derek{to} $1,000$. We set a lower peak learning rate \derek{to} $0.0002$ to prevent over-fitting. All the low-resource translation models are trained for 200 epochs. To prevent the models from over-fitting on low-resource language pairs, we set the dropout rate \derek{to} 0.3, both the attention dropout rate and activation dropout rate \derek{to} 0.1. For checkpoint selection, the checkpoints with the best validation BLEU would be selected as the final model checkpoints. We train an En-De model on the same parent data as the reversed parent model. We set $\tau$ \derek{to} 1.0 and tune $\alpha$ on $\{3,4,5,6,7\}$.
\paragraph{Evaluation} We used beam search with a beam width of 5 and a length penalty of 1 to evaluate all the NMT models. And we also use such decoding setting to generate pseudo parent source sentences. The translation quality is evaluated \derek{using} SacreBLEU \citep{DBLP:conf/wmt/Post18}\footnote{Signature: nrefs:1 + case:mixed + eff:no + tok:13a + smooth:exp + version:2.0.0} and \derek{the} BERTScore \citep{zhang2019BERTscore}\footnote{https://github.com/Tiiiger/bert\_score}, which evaluate the surface-form correctness and semantic correctness, respectively.
\begin{table*}[t]
    \centering
    \begin{tabular}{lrrrrrrrrrr}
    \toprule
\multirow{2}{*}{\bf Model} & \multicolumn{2}{c}{\bf Id-En} & \multicolumn{2}{c}{\bf Ca-En} &  \multicolumn{2}{c}{\bf Hu-En} & \multicolumn{2}{c}{\bf Pl-En} & \multicolumn{2}{c}{\bf Tr-En}  \\
\cmidrule(lr){2-3} \cmidrule(lr){4-5} \cmidrule(lr){6-7} \cmidrule(lr){8-9} \cmidrule(lr){10-11}
& \bf BLEU &\bf BS &\bf BLEU &\bf BS &\bf BLEU &\bf BS &\bf BLEU &\bf BS &\bf BLEU &\bf BS\\
      \midrule 
     Existing & 12.3 &-& 20.0 &- & 5.4 &- & 18.0 &- & 18.5 & -\\
      \hdashline
      Vanilla & 1.1 & 13.2& 1.1 & 15.5& 0.9 &0.9 & 1.5 & 18.9& 17.8 &51.8 \\
        ~~+ TL &  13.5& 37.7&  21.6&51.8 &  5.9& 27.4&19.9  & 55.3& 17.6&51.9 \\
        ~~+ TM-TL & 18.6 &49.9 & 25.3 & 58.9& 10.6 &41.2 & 21.4 & 58.2& 18.6&53.9 \\
             \hdashline
      ~~+ ConsistTL &\textbf{19.7} & \bf{52.2}&\textbf{26.6}&\bf{60.0} & \textbf{11.9}& \bf{43.9}& \textbf{22.4}& \bf{59.9}& \textbf{19.3} & \textbf{55.9}\\
    \bottomrule
    \end{tabular}
    \caption{Main results of the five translation tasks.   BS denotes the BERTScore. ``Existing'' are reported from \citet{DBLP:conf/emnlp/KhayrallahTPK20, DBLP:conf/emnlp/BaziotisHB20}. Both BLEU score and BERTScore reflect that ConsistTL achieves significant improvements compared with the strong baseline TM-TL on all the tasks ($p < 0.01$) \ \citep{koehn2004statistical}.}
    \label{tab:main}
\end{table*}
\subsection{Main Results}
Table \ref{tab:main} displays the results of the five aforementioned translation tasks \derek{in terms of the} BLEU \derek{score} and \derek{the} BERTScore.
We report the existing \derek{results} from \citet{DBLP:conf/emnlp/KhayrallahTPK20} and \citet{DBLP:conf/emnlp/BaziotisHB20}. 
ConsistTL outperforms the existing works on the same test sets in terms of \derek{the} BLEU score.
\paragraph{Global Voices \{Id,Ca,Hu,Pl\}-En}
The data scale of the four language pairs (Id-En, Ca-En, Hu-En, Pl-En) collected from Global Voices is extremely lower than \derek{that of} WMT17 Tr-En. The parameter transfer methods TL and TM-TL are strong since they outperform vanilla NMT in terms of both BLEU and BERTScore. Our method ConsistTL achieves a significant average improvement of 1.1 BLEU and 2.0 BERTScore over the TM-TL.

\paragraph{WMT17 Tr-En}
Different from the four aforementioned language pairs, the data scale of Tr-En is larger. Compared with the aforementioned extremely low-resource language pairs, the transfer learning baseline methods \derek{have lower} performance gains (TM-TL) or even no \derek{performance} gains (TL) over Vanilla NMT on this benchmark with larger data. Although baseline methods are less informative on the Tr-En, our method still achieves a significant improvement over the strong baseline TM-TL with 0.7 BLEU and 2.0 BERTScore. The results of BERTScore on five translation benchmarks indicate that ConsistTL \derek{provides} consistent and significant improvements \derek{of the} semantic correctness. Our follow-up analysis is based on the WMT17 Tr-En benchmark.
\section{Analysis}
\subsection{Effect of Loss Type}
\begin{table}[t]
    \centering
    \begin{tabular}{llll}
    \toprule
    \textbf{Tr-En} & \textbf{Consistency Loss} & \textbf{Valid}&\textbf{Test}\\
    \midrule
        TL &- & 17.8&17.6 \\
        TM-TL &- & 18.8 & 18.6 \\ 
        \hdashline
        \multirow{2}{*}{ConsistTL}  & Kullback-Leibler & 19.2 &18.9 \\
                  &Jensen–Shannon & 19.8 &19.3\\
    \bottomrule
    \end{tabular}
    \caption{Comparison of different consistency loss. ConsistTL works better with JS loss than KL loss.}
    \label{tab:measure-function}
\end{table}
There are many ways to measure the difference between two distributions. Different measure functions would result in different consistency loss types. This part compares two typical choices {Jensen–Shannon divergence \citep{DBLP:journals/tit/Lin91} and Kullback–Leibler divergence \citep{kullback1951information}} to study the effect of the chosen loss type \derek{on} ConsistTL. To make the most of their capabilities, we tune the hyperparameters $(\alpha,\tau)$ for \derek{each} loss type. The final hyperparameters of \derek{the} JS loss and \derek{the} KL loss are $(7.0, 1.0)$ and $(4.0,2.0)$ respectively. As seen in Table \ref{tab:measure-function}, the model with JS achieves the best improvements, while the model with KL loss realizes limited improvements. Perhaps minimizing \derek{the} KL divergence is more difficult than \derek{minimizing the} JS divergence since we observe that lowering \derek{the} KL loss \derek{through} smoothing ($\tau > 1$) is helpful.
\subsection{Comparison of Learning Curves}
\begin{figure}[t]
    \centering
    \includegraphics[width=0.85\columnwidth]{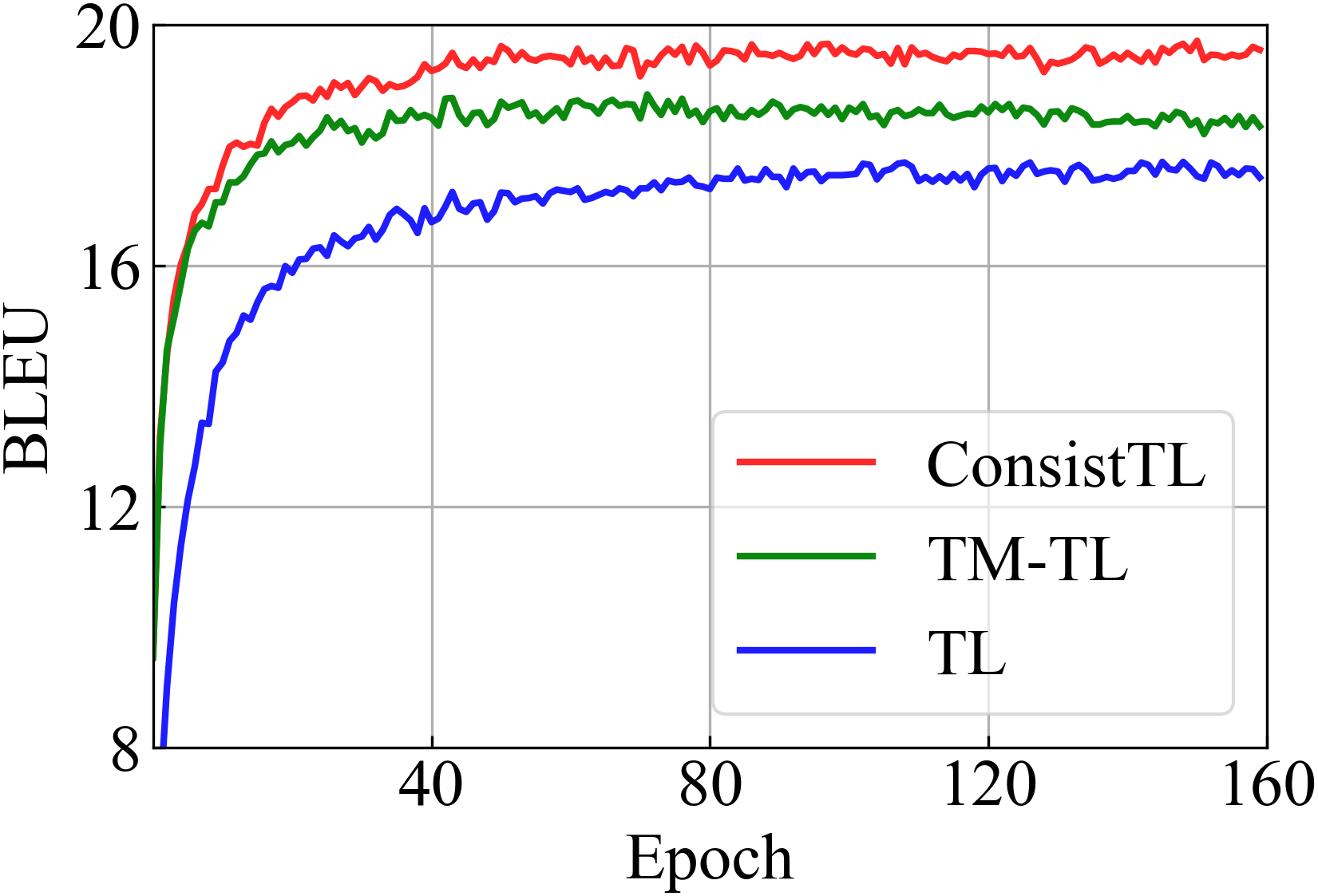}
    \caption{Learning curves of different TL methods. ConsistTL achieves the largest upper bound among them.}
    \label{fig:learning curve}
\end{figure}

The learning curve is one way to describe the temporal effect of the proposed method during training\derek{, which has also been} adopted in previous consistency modeling works \citep{DBLP:conf/nips/LiangWLWMQCZL21,DBLP:conf/acl/KambhatlaBS22, xie2021target}. 
This part compares the proposed method\derek{,} ConsistTL\derek{,} with the baseline methods TL and TM-TL \derek{in terms of} their learning curves described by validation set BLEU. Figure \ref{fig:learning curve} displays the learning curves of these three transfer learning methods. During the child training, the ConsistTL and TM-TL grow into larger validation BLEU values with higher convergence speeds. The difference between ConsistTL and TM-TL is invisible in the early learning phase and gradually changes into a stable and significant value. Finally, ConsistTL achieves the largest upper bound among these three learning curves, which \derek{indicates} that ConsistTL is informative to child training.
\subsection{Effect of Pseudo Source Generation}
\begin{table}[t]
    \centering
    \begin{tabular}{llll}
    \toprule
    \textbf{Tr-En} & \textbf{Parent Source} & \textbf{Valid}&\textbf{Test}\\
    \midrule
        TL &- & 17.8&17.6 \\
        TM-TL &- & 18.8 & 18.6 \\ 
        \hdashline
        \multirow{3}{*}{ConsistTL}  & Sampling & 19.3 &18.8 \\
        &Greedy Search & 19.8 &19.3  \\ 
                  &Beam Search & 19.8 &19.3 \\
    \bottomrule
    \end{tabular}
    \caption{Effect of pseudo parent source generation. Both greedy search and beam search can ensure the semantic equivalence of pseudo parent sources.}
    \label{tab:pseudo-parent-source}
\end{table}
Although beam search is a de facto decoding method \derek{that is used} to generate sentences in the NMT task, there are also other decoding methods \derek{that can be used } to generate synthetic data \citep{ott2018analyzing,DBLP:conf/emnlp/EdunovOAG18}, including greedy search, and greedy sampling. To study the effect of decoding methods in parent-side synthetic data generation, we compare the results produced by beam search, greedy search and greedy sampling. \derek{As shown} in Table \ref{tab:pseudo-parent-source}, we find that greedy sampling produces lower improvement in contrast to the beam search. This can be interpreted \derek{to mean} that greedy sampling introduces low-quality translations due to the randomness during decoding \citep{DBLP:conf/acl/IppolitoKSKC19}. Though greedy search performs worse than beam search in usual model testing, {the final results} in Table \ref{tab:pseudo-parent-source} empirically shows that there is no significant difference in final results between beam search and greedy search. That means the translation quality of greedy search meets the ConsistTL requirement. Therefore, we can adopt the greedy search to generate pseudo parent source sentences for future usage.
\subsection{Effect of Data Scale}
\begin{figure}[t]
    \centering
    \includegraphics[width=0.8\columnwidth]{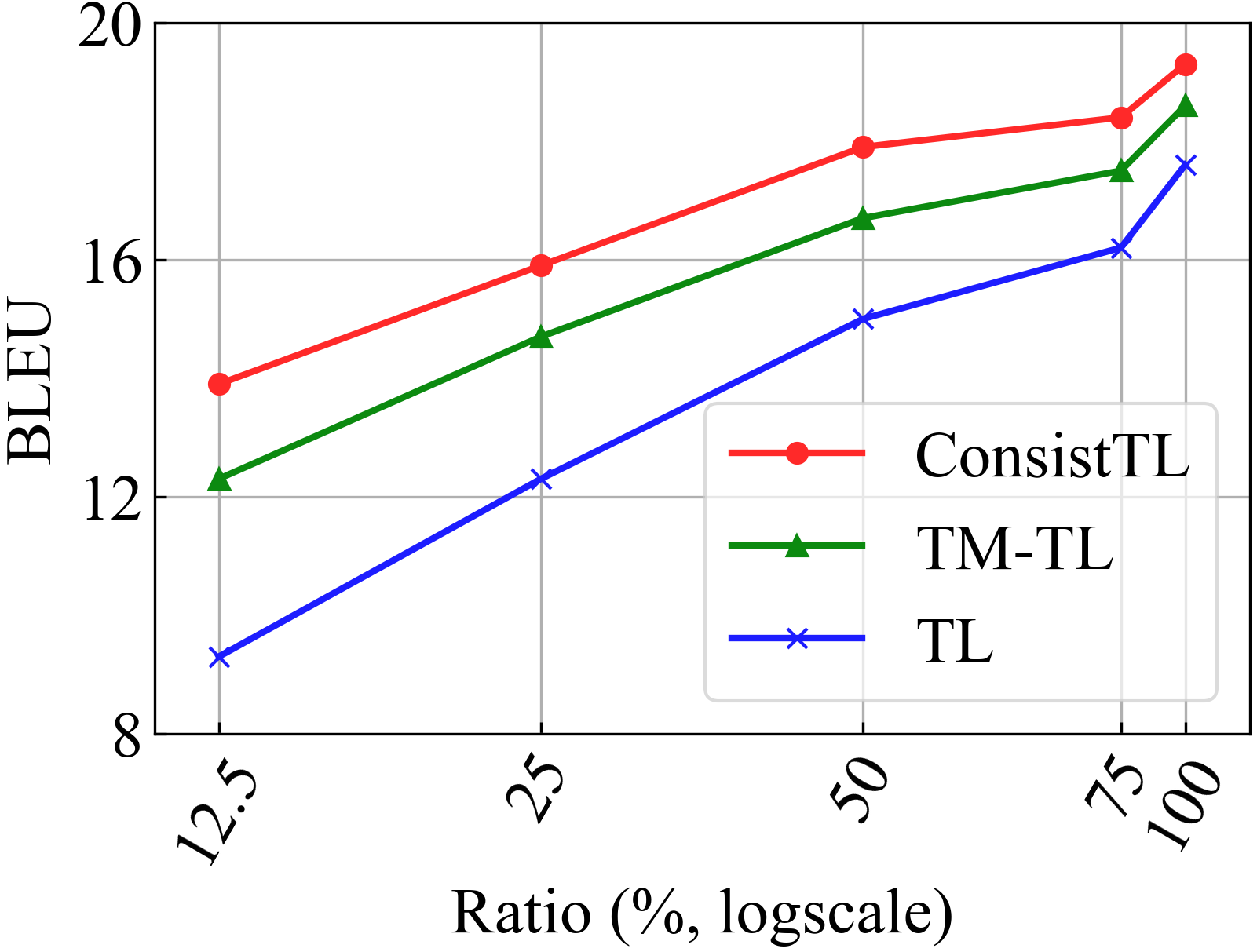}
    \caption{BLEU score {with varying data scale.}}
    \label{fig:data_scale}
\end{figure}
To verify the robustness of the improvements brought by ConsistTL across data scales, we compare our method and {the baselines methods, TM-TL and TL, on the subsets} with different scales. The training subsets are randomly sampled from the full Tr-En training data according to the specific ratios. Figure \ref{fig:data_scale} demonstrates the result of data ablation. The performances of ConsistTL and TM-TL are more stable than TL on different \derek{data scales, }since TL drops more performance on smaller-scale data. ConsistTL achieves significant improvements consistently in contrast to TM-TL across different data scales. This result indicates the robust effectiveness of ConsistTL with data scales varying.
\subsection{Effect of Back-Translation}
\begin{table}[t]
    \centering
    \begin{tabular}{lcll}
    \toprule
    \textbf{Tr-En} & \textbf{w/ BT} & \textbf{Valid}&\textbf{Test}\\
    \midrule
        {TM-TL~\scriptsize{\citep{DBLP:conf/acl/AjiBHS20}}} &Y & - & 20.6 \\ 
                \hdashline
        \multirow{2}{*}{TM-TL (Ours)}  & N & 18.8 &18.6 \\
                  &Y & 22.3 &21.6 \\
        \hdashline
        \multirow{2}{*}{ConsistTL}  & N & 19.8 &19.3 \\
                  &Y & 22.9 &22.3\\
    \bottomrule
    \end{tabular}
    \caption{Effect of combining with back-translation. ConsistTL still \derek{outperforms} TM-TL+BT significantly.}
    \label{tab:back-translation}
\end{table}
Back-translation \citep{DBLP:conf/acl/SennrichHB16,DBLP:conf/emnlp/EdunovOAG18} is an essential method for NMT \derek{that is used} to augment training corpus by generating synthetic sentences from target-side monolingual data. In this part, to study the complementarity between this work and back-translation, we compare the proposed method and the baseline TM-TL on the training data augmented by back-translation. We sample 200k English monolingual data from News Crawl 2015 
for back-translation, following the approximate 1:1 ratio according to \citet{DBLP:conf/acl/AjiBHS20}. 
{In \citet{DBLP:conf/acl/AjiBHS20}, the parent model for Tr-En is also a Transformer-base model trained on WMT17 De-En data.} Table \ref{tab:back-translation} displays the results reported from \citet{DBLP:conf/acl/AjiBHS20} and the results of our implementation. Our proposed method ConsistTL outperforms the existing baselines with 1.7 BLEU. In combination with back-translation, our method yields a performance gain with the significance of 95\% ($p<0.01$) on the test set, which means ConsistTL is complementary to back-translation. {This result confirms the generality of ConsistTL since ConsistTL can be generalized to a scenario of \derek{in which} training data \derek{are} augmented by back-translation.}
\subsection{Combination with Inner Consistency}
\begin{table}[t]
    \centering
    \begin{tabular}{lcll}
    \toprule
    \textbf{Tr-En} &\bf w/ Inner Consist. & \textbf{Valid} & \textbf{Test}\\
    \hline
        \multirow{2}{*}{TM-TL} &N & 18.8 & 18.6 \\
        &Y & 20.1 & 20.0 \\ \hdashline
       \multirow{2}{*}{ConsistTL} &N & 19.8 & 19.3 \\
       &Y & 20.8 & 20.6 \\
     \hline
    \end{tabular}
    \caption{Effect of combining with inner consistency modeling.}
    \label{tab:Inner-Consistency}
\end{table}
As ConsistTL learns the cross-model consistency, here we investigate the effectiveness of such cross-model consistency under inner consistency modeling. \lzc{R-Drop is a typical method to model inner consistency} \citep{DBLP:conf/nips/LiangWLWMQCZL21}. We compare the transfer learning methods combined with inner consistency implemented by R-Drop. Table \ref{tab:Inner-Consistency} shows that our method achieves the best performance and consistent improvements across the validation set and test set ($p<0.01$). This result validates that the cross-model consistency is indeed effective since ConsistTL still improves performance when combined with R-Drop.
\subsection{Model Calibration}
\begin{figure}[t]
    \centering
    \includegraphics[width=0.85\columnwidth]{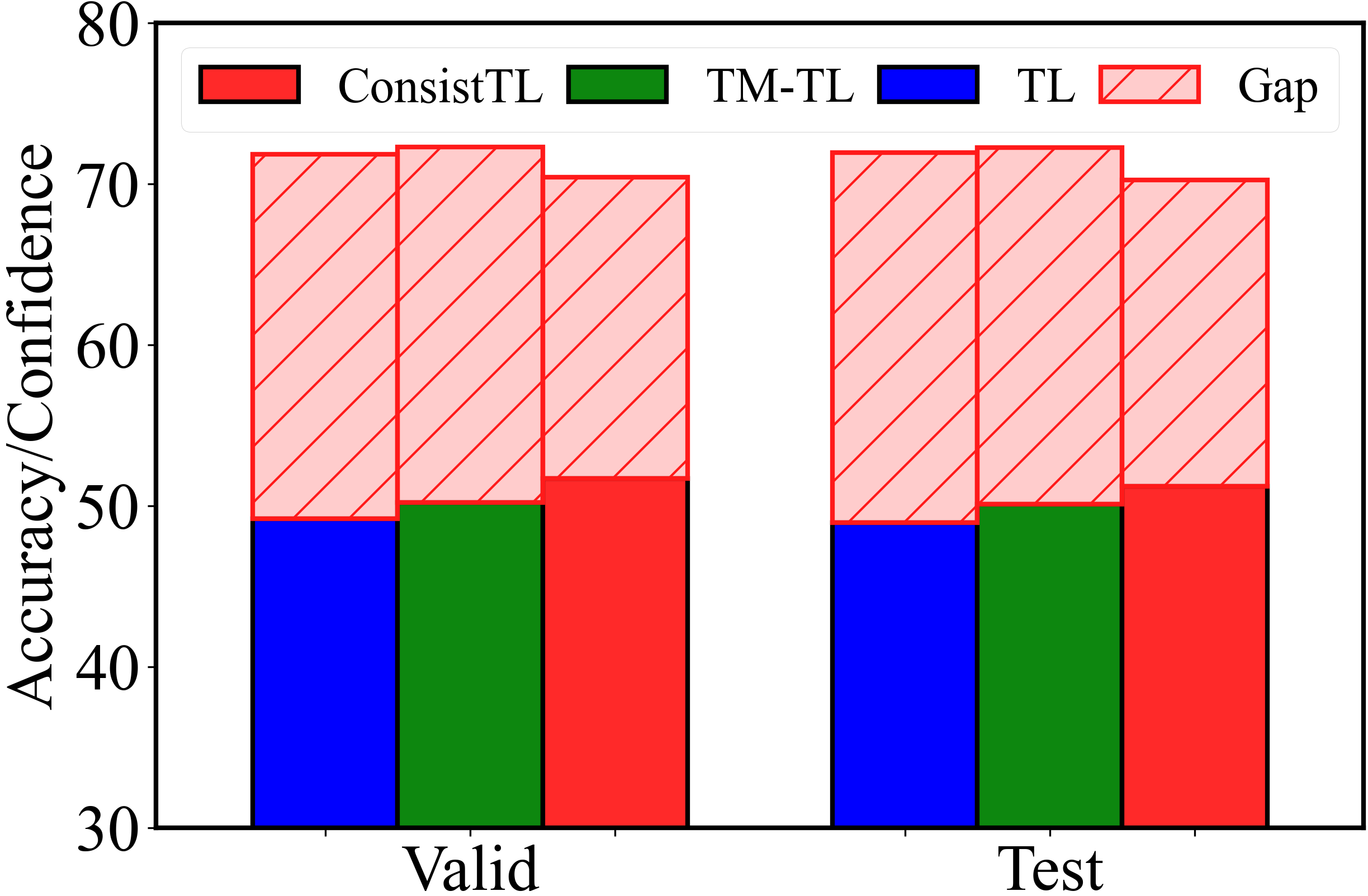}
    \caption{Visualization of prediction confidence and accuracy for the Tr-En model. The complete column represents the confidence while the pure color column represents the accuracy. The striped part denotes the gap between confidence and accuracy. ConsistTL narrows the gap and achieves the best prediction accuracy, resulting in better model calibration.}
    \label{fig:InfECE}
\end{figure}

As this paper proposes leverag\derek{ing} the prediction distribution of parent model, \derek{the following questions arises}: {\textit{how does \derek{the} parent prediction help shape \derek{the} child prediction?}} The inference calibration \derek{can be used} to describe the prediction distribution for NMT models, which requires mitigating the gap between \derek{the} prediction confidence and translation accuracy during inference \citep{DBLP:conf/acl/WangTSL20}. 
We exploit the gap between the averaged confidence and translation accuracy to evaluate inference calibration\footnote{https://github.com/shuo-git/InfECE}. The average accuracy illustrates the correctness of generated tokens, which is annotated by \derek{the} translation error rate (TER)~\citep{snover-etal-2006-study}. 
The averaged confidence is computed on the estimated probabilities of generated tokens.
More details can refer to~\citet{DBLP:conf/acl/WangTSL20}.
Figure \ref{fig:InfECE} shows the averaged accuracy and the averaged confidence of child models implemented by ConsistTL, TM-TL, and TL. 
The visualization results show that \lzc{the averaged confidences are higher than the corresponding averaged accuracy, namely overconfidence of the models.} 
Compared with other methods, ConsistTL improves \derek{the} translation accuracy and reduces overconfidence.
In this way, ConsistTL narrows the gap between \derek{the} confidence and accuracy, which indicates that ConsistTL can be beneficial to child model inference calibration by introducing the prediction distribution of \derek{the} parent model. This \derek{provides} a reasonable explanation \derek{of} why ConsistTL can \derek{give a} performance boost to \derek{the} child model.

\section{Conclusion and Future Work}
We introduce ConsistTL for transfer learning in low-resource NMT, which can continuously transfer the knowledge from parent prediction during child training.
In order to transfer knowledge from parent prediction, we propose to model cross-model consistency between parent model and child model with the child source sentences and pseudo parent source sentences. 
Experimental results on five translation benchmarks verify the effectiveness of ConsistTL, which can significantly improve the translation performance.
ConsistTL is \lzc{complementary} to other powerful methods for low-resource NMT (e.g., back-translation).
Further analysis reveals that introducing parent prediction is helpful to shape child model prediction distribution, resulting in better inference calibration.

In the future, we would like to apply curriculum learning~\citep{liu-etal-2020-norm,zhan2021meta} to better organize the learning of the child model. It is also worthwhile to enhance the parent and child models by utilizing pre-trained knowledge learned from unlabeled data~\cite{liu-etal-2021-complementarity-pre, liu-etal-2021-copying}.
\section*{Limitations}
Our proposed framework would occupy extra computation resources compared with baseline methods. 
In the part of semantically-equivalent parent data construction, we need to train an additional reversed parent model to back-translate the target sentences of child data.
In the part of parent-child cross-model consistency learning, at each training step of child model, the parent model \lzc{needs} an additional forward pass to generate prediction distribution for guidance.
\section*{Acknowledgments}
This work was supported in part by the National Natural Science Foundation of China (Grant No. 62206076), the Science and Technology Development Fund, Macau SAR (Grant No. 0101/2019/A2), Shenzhen College Stability Support Plan (Grant No. GXWD20220811173340003 and GXWD20220817123150002) and the Multi-year Research Grant from the University of Macau (Grant No. MYRG2020-00054-FST). 
We would like to thank the anonymous reviewers and meta-reviewer for their insightful suggestions.
\bibliography{emnlp2022}
\bibliographystyle{acl_natbib}

\end{document}